\documentclass{article}

\usepackage{arxiv}

\usepackage[utf8]{inputenc} 
\usepackage[T1]{fontenc}    
\usepackage{hyperref}       
\usepackage{url}            
\usepackage{booktabs}       
\usepackage{amsfonts}       
\usepackage{nicefrac}       
\usepackage{microtype}      
\usepackage{lipsum}		
\usepackage{graphicx}
\usepackage{doi}

\usepackage{xcolor}
\usepackage{amsmath}

\newcommand{\dahntab}[1]{
  \newbox\mybok%
  \setbox\mybok=\hbox{\vbox{
      \begin{tabbing}
        #1
      \end{tabbing}%
    }}

  \newdimen\bokwidth%
  \bokwidth=\wd\mybok%
  \newdimen\myl%
  \myl=\textwidth%
  \divide\myl by 2%
  \divide\bokwidth by -2%
  \advance\myl by\bokwidth%
  \vrule width\myl height 0pt depth 0pt%
  \usebox\mybok%
}

\usepackage{booktabs, makecell, tabularx}

\usepackage{siunitx}%
\usepackage{adjustbox}
\usepackage{algorithm} 
\usepackage{algorithmic}  
\usepackage[algo2e]{algorithm2e}

\SetCommentSty{mycommfont}

\SetKwInput{KwInput}{Input}                
\SetKwInput{KwOutput}{Output}              
\SetKwInOut{Parameter}{Parameters}

\title{An automated threshold Edge Drawing algorithm}

\author{ \href{https://orcid.org/0000-0002-0071-958X}{\includegraphics[scale=0.06]{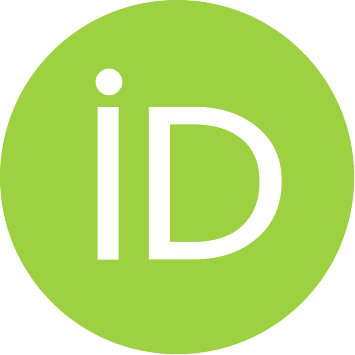}\hspace{1mm}Ciprian~Orhei}\\
	Politehnica University of Timi\c{s}oara\\
	Timi\c{s}oara, Romania\\
	\texttt{ciprian.orhei@cm.upt.ro} \\
	\And
	\href{https://orcid.org/0000-0002-0203-1519}{\includegraphics[scale=0.06]{orcid.pdf}\hspace{1mm}Muguras~Mocofan} \\
	Politehnica University of Timi\c{s}oara\\
	Timi\c{s}oara, Romania\\
	\texttt{muguras.mocofan@upt.ro} \\
	\And
	\href{https://orcid.org/0000-0003-2394-4859}{\includegraphics[scale=0.06]{orcid.pdf}\hspace{1mm}Silviu~Vert} \\
	Politehnica University of Timi\c{s}oara\\
	Timi\c{s}oara, Romania\\
	\texttt{silviu.vert@upt.ro} \\
	\And
	\href{https://orcid.org/0000-0003-1185-1997}{\includegraphics[scale=0.06]{orcid.pdf}\hspace{1mm}Radu~Vasiu} \\
	Politehnica University of Timi\c{s}oara\\
	Timi\c{s}oara, Romania\\
	\texttt{radu.vasiu@upt.ro} \\
}

\renewcommand{\shorttitle}{Accepted in 44th International Conference on Telecommunications and Signal Processing (TSP)}

\hypersetup{
pdftitle={An automated threshold Edge Drawing algorithm},
pdfsubject={cs.CV},
pdfauthor={Ciprian~Orhei, Muguras~Mocofan, Silviu~Vert, Radu~Vasiu},
pdfkeywords={Edge detection, Edge Drawing (ED), Otsu threshold, Automated threshold},
}

\begin{document}
\maketitle

\begin{abstract}
Parameter choosing in classical edge detection algorithms can be a costly and complex task. Choosing the correct parameters can improve considerably the resulting edge-map. In this paper we present a version of Edge Drawing algorithm in which we include an automated threshold choosing step. To better highlight the effect of this additional step we use different first order operators in the algorithm. Visual and statistical results are presented to sustain the benefits of the proposed automated threshold scheme.
\end{abstract}

\keywords{Edge detection, Edge Drawing (ED), Otsu threshold, Automated threshold}

\section{Introduction}

Extracting edge features from images is a problem that was tackled in many ways by researchers. The proposed solutions vary from classical Sobel  \cite{1973SobelOriginal} or Prewitt \cite{1970PrewittOriginal} 
to more complex algorithms like Canny \cite{canny86} or Edge Drawing \cite{EdgeDrawing2012}. Edges remain a basic yet important feature in the Computer Vision domain.

Edge Drawing -ED- proposes an edge segment detection algorithm that applies the high-level cognitive reasoning employed in dot-to-dot boundary completion puzzles to edge segment detection. First, we analyze the effect on the original ED algorithm when we use other first order derivative operators. Afterwards, we propose a scheme for selecting the threshold parameters that is dependent on the image. 

For an algorithm such as ED algorithm, in which we can configure $4$ parameters (gradient threshold, anchor threshold, scan interval and Gaussian kernel size), we can easily reach over $100$ variants to fine tune it on a dataset. The fine-tuning phase is dependent on the image set and use case. From experience, the setting of parameters can become a lot more facile but it still remains a costly process.

Extensions of the ED algorithm exist in literature where additional steps are included, such as: edge segments are validated by an "a contrario" step due to the Helmholtz principle \cite{akinlar2012edpf} or edge segments are linked based on the predictions generated from past movements \cite{akinlar2016pel}. But for this research we attempt to maintain the real-time benefit of the original algorithm and adopt a less costly additional step. In this direction, we propose a scheme of choosing the ED threshold that is not user dependent but contextual of the image. The Otsu threshold in our opinion is a good starting point to adapt the necessary thresholds of the algorithm. Automated threshold selection phases were added to other edge detection algorithm as we can see in \cite{otsu_impr_1, otsu_impr_4, otsu_impr_6}.

We consider that an automated threshold choosing algorithm is a important improvement considering the dependencies between the data we process and the gradients obtained in the processing phase.

The paper is organized as follows: in Section \ref{Section:Preliminaries}, we present the concepts needed as background for our modifications and simulation flow; in Section \ref{Section:ED_proposed}, the original and proposed ED algorithms are described; in Section \ref{Section:ED_results}, the results of our simulations are analyzed.

\section{Research background}
\label{Section:Preliminaries}

\subsection{First Order derivative Operators}

For our experiments, we use the following discrete differentiation operators: the \textbf{Sobel Operator} \cite{1973SobelOriginal}, the \textbf{Prewitt Operator} \cite{PrewittOriginal1970}, the \textbf{Kirsch operator} \cite{1971Kirschcomputer} , the \textbf{Kitchen Operator}\cite{Kitchen1989}, the \textbf{Kayalli Operator} \cite{2000Kayyali}, the \textbf{Scharr Operator} \cite{2000ScharrOriginal}, the \textbf{Kroon Operator} \cite{KroonOriginal2009}, and the \textbf{Orhei Operator} \cite{2020Orhei}. All the used kernels are presented in Figure \ref{figure:3_kernel_masks}.

\begin{figure}[h!]
\centering
\small{
\setlength{\tabcolsep}{0.5pt}

\begin{tabular}{cccc}
    $\begin{bmatrix}
        -1 & 0 & 1 \\
        -2 & 0 & 2 \\
        -1 & 0 & 1 \\
    \end{bmatrix}$
    &
    $\begin{bmatrix}
        -1 & 0 & 1 \\
        -1 & 0 & 1 \\
        -1 & 0 & 1 \\
    \end{bmatrix}$
    &
    $\begin{bmatrix}
        -3 & -3 & 5 \\
        -3 &  0 & 5 \\
        -3 & -3 & 5 \\
    \end{bmatrix}$
    &
    $\begin{bmatrix}
         -2 & 0 & 2 \\
         -3 & 0 & 3 \\
         -2 & 0 & 2 \\
    \end{bmatrix}$
    \\
    ~
    \\
        Sobel 
    &         
        Prewitt
    &
        Kirsch
    &
        Kitchen
    
    \\
     ~
    \\
    $\begin{bmatrix}
         -6 & 0 &  6 \\
          0 & 0 &  0 \\
          6 & 0 & -6 \\
    \end{bmatrix}$
    &
    $\begin{bmatrix}
         -3 & 0 &  3 \\
        -10 & 0 & 10 \\
         -3 & 0 &  3 \\
    \end{bmatrix}$
    &
    $\begin{bmatrix}
        -17 & 0 & 17 \\
        -61 & 0 & 61 \\
        -17 & 0 & 17 \\
    \end{bmatrix}$
    &
    $\begin{bmatrix}
        -1 & 0 & 1 \\
        -4 & 0 & 4 \\
        -1 & 0 & 1 \\
    \end{bmatrix}$
    \\
    ~
    \\
        Kayyali
    &
        Scharr
    &
        Kroon
    &
        Orhei
    \\ 
\end{tabular}
}
\caption{First Order derivative edge operators kernels}
\label{figure:3_kernel_masks}
\end{figure}

The gradient is a measure of change in a function, and an image can be considered to be an array of samples of some continuous function of image intensity, typically  two-dimensional equivalent of first derivative. Gradient magnitude is calculated using Formula \ref{formula:magnitude_and_aprox}, where $f(x,y)$ is the image and $G_x$, $G_y$ are the components on $x$ and $y$ axis \cite{haralick1992computer}.

\begin{align}\label{formula:magnitude_and_aprox}
    G[f(x,y)] = \sqrt{G_x^2 + G_y^2} \approx |G_x| + |G_y|
\end{align}

\subsection{Otsu thresholding}

Otsu’s thresholding method corresponds to the linear discriminant criteria, which assumes that the image consists of only two objects: foreground and background \cite{otsu}. Otsu’s method determines the threshold value based on the statistical information of the image where the variance of clusters $T_0$ and $T_1$ can be computed. Otsu’s algorithm tries to find a threshold value (t) which minimizes the weighted within-class variance given by the relation observed in Equations \ref{formula:otsu}.

\begin{align}\label{formula:otsu}
    \sigma_{\sigma}^{2}(t) = \omega_{0}(t) * \sigma_{0}^{2}(t) + \omega_{1}(t) * \sigma_{1}^{2}(t) \\ = \omega_{0}(t) * \omega_{1}(t){[\mu_{0}(t)-\mu_{1}(t)]}^2
\end{align}

Weights $\omega_{0}$ and $\omega_{1}$ are the probabilities of the two classes separated by a threshold $t$ and $\sigma_{0}^{2}$ and $\sigma_{1}^{2}$ are variances of these two classes. The class probability $\omega_{0,1}(t)$ is computed from the $L$ bins of the histogram. 

For two classes, minimizing the intra-class variance is equivalent to maximizing inter-class variance as we can see in Equation \ref{formula:otsu}, which is expressed in terms of class probabilities $\omega$ and class means $\mu$, where the class means $\mu _{0}(t)$, $\mu _{1}(t)$ and $\mu _{T}$ are presented in Equation \ref{formula:otsu_class_mean_0} - \ref{formula:otsu_class_mean_T}.

\begin{align}\label{formula:otsu_class_mean_0}
    \mu_{0}(t) = \frac{\sum_{i=0}^{t-1}i*p(i)}{\omega_0(t)}
\end{align}

\begin{align}\label{formula:otsu_class_mean_1}
    \mu_{1}(t) = \frac{\sum_{i=t}^{L-1}i*p(i)}{\omega_1(t)}
\end{align}

\begin{align}\label{formula:otsu_class_mean_T}
    \mu_{T}(t) = \sum_{i=0}^{L-1}i*p(i)
\end{align}




\subsection{Benchmarking the edge operators}

For highlighting the results obtained, we use BSDS500 \cite{Bsds2011} which contains a dataset of natural images that have been manually segmented. The human annotations serve as ground truth for the  benchmark for comparing different segmentation and boundary detection algorithms. For evaluating the images  generated from algorithms to the ground truth images, the Corresponding Pixel Metric (CPM)  algorithm \cite{CPM2003} is used.

For each image, two quantities Precision ($P$) and Recall ($R$) will be computed, as were defined in \cite{F12007}. Precision, with Formula \ref{precision}, represents the probability that a resulting edge/boundary pixel is a true edge/boundary pixel. Recall, with Formula \ref{recall}, represents the probability that a true edge/boundary pixel is detected. In these formulas, $TP$ (True Positive) represents the number of matched edge pixel, $FP$ (False Positive) the number of edge pixels which are incorrectly highlighted and $FN$ (False Negative) the number of pixel that have not been detected. Those two quantities are used to compute \textit{F-measure} (F1-score) by applying the Formula \ref{f-measure}.

\begin{align}\label{precision}
    P = \frac{TP}{TP+FP}.
\end{align}

\begin{align}\label{recall}
    R = \frac{TP}{TP+FN}.
\end{align}


\begin{align}\label{f-measure}
F-measure = \frac{2*TP}{2*TP+FP+FN}.
\end{align}

\section{Proposed Edge Drawing algorithm}
\label{Section:ED_proposed}

\begin{algorithm}[H]
    \KwIn{$image$}
    \Parameter{$gaussian\_kernel\_size$}
    \KwOut{edge segments}    

    \tcc{ED algorithm works on grey-scale images}
    $image \leftarrow$ Convert image to grey-scale

    \tcc{Gaussian filter with $gaussian\_kernel\_size$ param}
    $image \leftarrow$ Apply Gaussian filter smoothing
    
    \tcc{Kernels from Fig. \ref{figure:3_kernel_masks} using Formula \ref{formula:magnitude_and_aprox}}
    $grad\_map \leftarrow$ Calculate gradient map
    
    \tcc{$|G_x| \geq |G_y|$ vertical edge else horizontal edge}
    $orientation\_map \leftarrow$ Calculate direction map
    
    \tcc{Find Otsu Threshold for 2 classes}
    $thr\_otsu \leftarrow$ Apply Otsu transformation on $image$
    
    \tcc{Find threshold according to Equation \ref{formula:ed_grad} and \ref{formula:anc_grad}}
    $grad\_thr, anchor\_thr \leftarrow$ Calculate thresholds for ED
    
    \tcc{Apply global threshold scheme by $grad\_thr$}
    $grad\_map \leftarrow$ Threshold gradient map

    \tcc{Find anchors using $anchor\_thr$ and $scan\_interval$}
    $anchor\_list \leftarrow$  Extract anchors
    
    \tcc{Three immediate neighbors are considered}
    $edge\_segments \leftarrow$ Smart routing
\caption{Modified Edge Drawing Algorithm}
\label{algorithm:ed_mod}
\end{algorithm}

The ED edge detection algorithm presented good results for a traditional edge detection concept. But we are concerned with the amount of variants we need to test for fine tuning the parameters for a given dataset or scenario. 

ED proposes an edge segment detection algorithm that applies the high-level cognitive reasoning employed in dot-to-dot boundary completion puzzles to edge segment detection \cite{EdgeDrawing2012}.

\begin{figure*}[!h]
    \begin{minipage}{0.3\textwidth}
        \centering
        \includegraphics[scale=0.17]{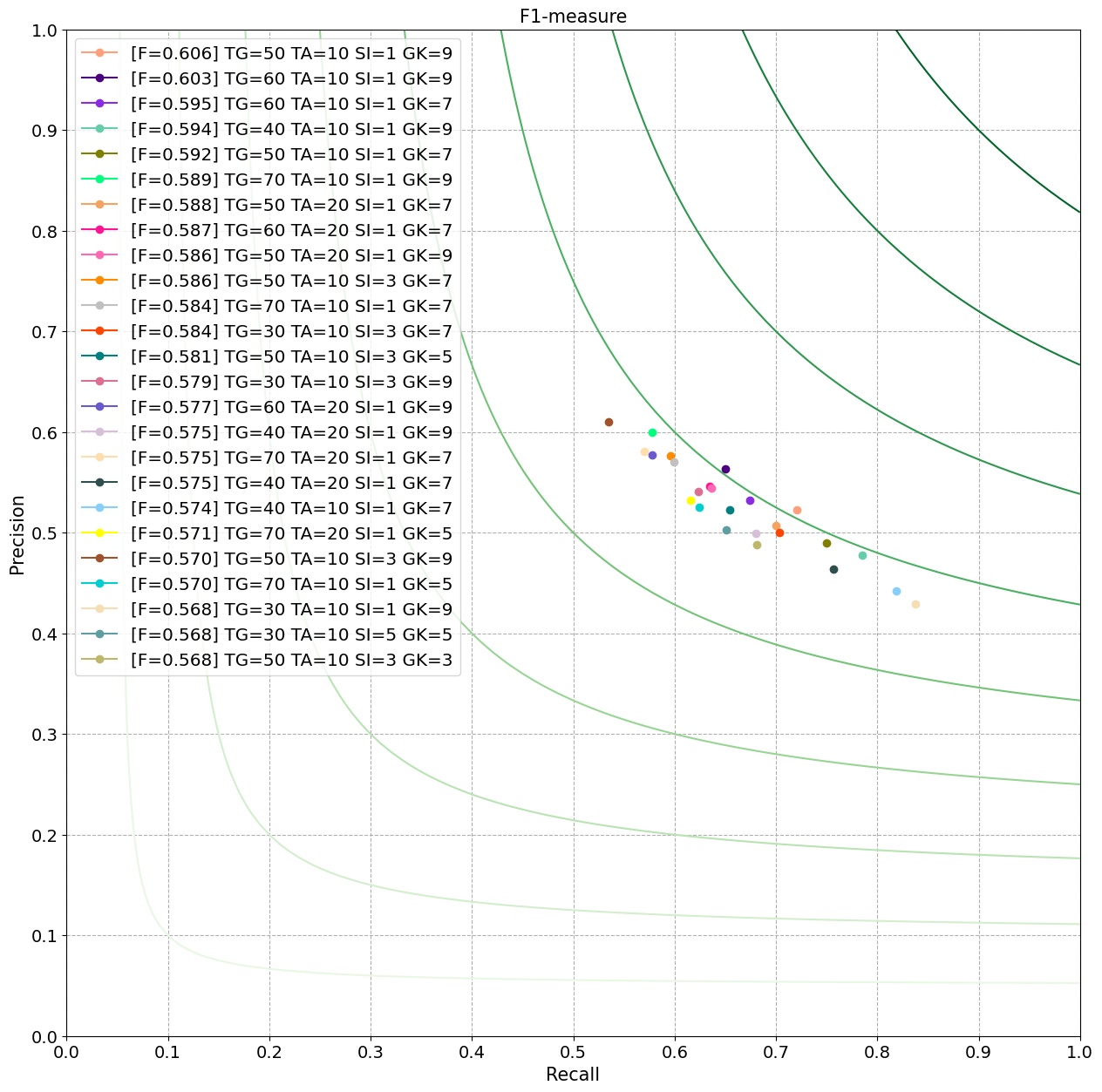} 
        \caption{ED parameter tuning}
        \label{fig:sobel_fine_tuning_natural}
    \end{minipage}\hfill
    \begin{minipage}{0.3\textwidth}
        \centering
        \includegraphics[scale=0.17]{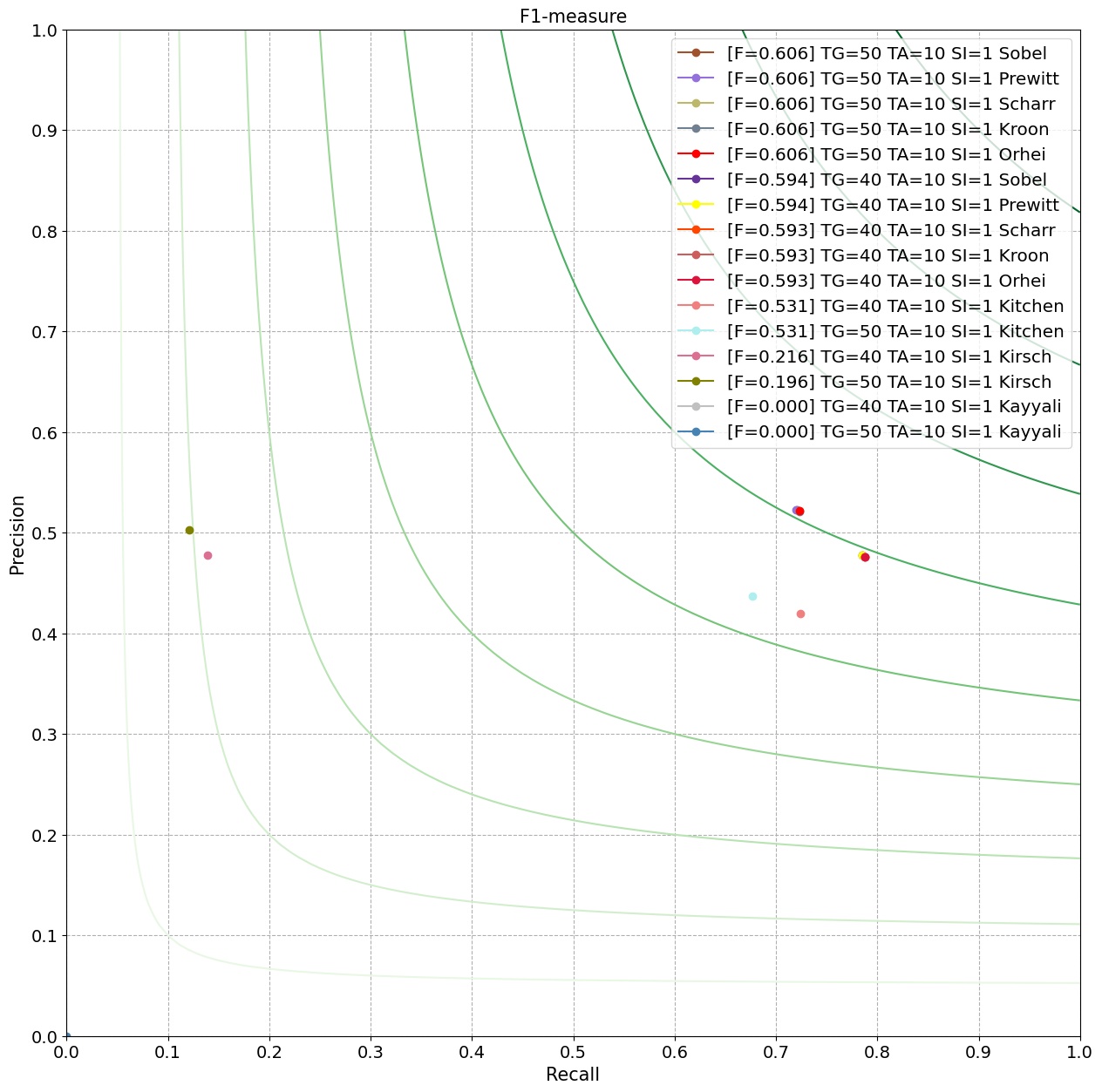} 
        \caption{ED results using different operators}
        \label{fig:ed_natural}
    \end{minipage}\hfill
    \begin{minipage}{0.3\textwidth}
        \centering
        \includegraphics[scale=0.17]{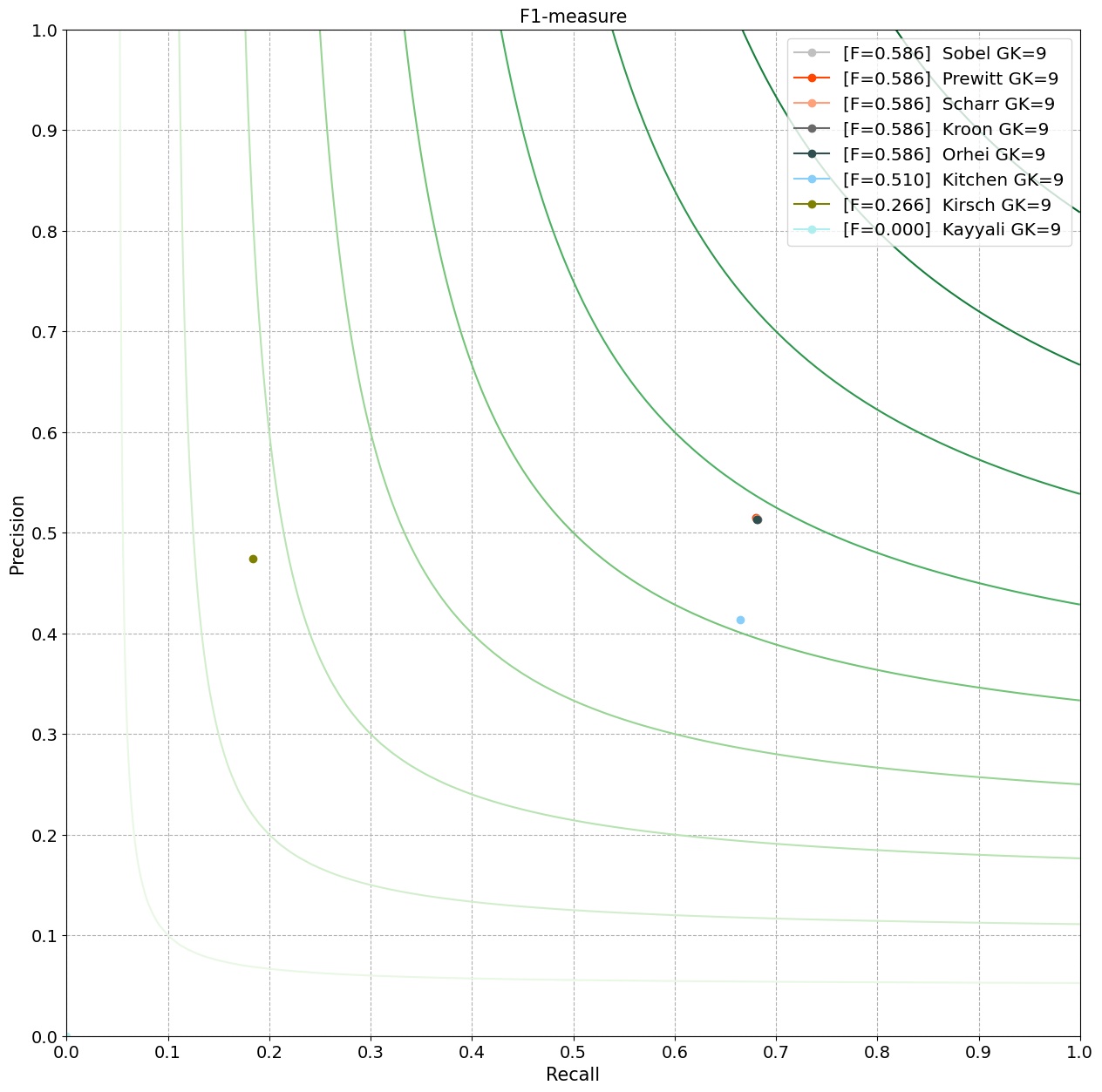} 
        \caption{ED proposed results}
        \label{fig:ed_proposed}
    \end{minipage}\hfill
\end{figure*}

The original ED algorithm consists of the following steps: first the image is smoothed using a Gaussian filter \cite{haralick1992computer}; afterwards the horizontal and vertical gradients are calculated using Formula \ref{formula:magnitude_and_aprox}; the edge direction map is calculated by using the idea that $|G_x| \geq |G_y|$; suppression of the so called "weak" pixels is done and anchor points are extracted \cite{EdgeDrawing2012}. 

Anchors would be located at the peaks of the gradient map. To connect consecutive anchors, we simply go from one anchor to the next by proceeding over the cordillera peak of the gradient map mountain. This process is guided by the gradient magnitude and edge direction maps computed \cite{EdgeDrawing2012}.

Otsu is an approach that separates the image into background and foreground, which makes a perfect candidate to take in consideration for choosing our thresholds. In this scenario, we propose that for an image we find the Otsu threshold and afterwards we choose the gradient threshold and anchor threshold accordingly to Equation \ref{formula:ed_grad} and Equation \ref{formula:anc_grad}.

\begin{align}\label{formula:ed_grad}
    T_{grad} = 0.5 * T_{otsu}
\end{align}

\begin{align}\label{formula:anc_grad}
    T_{anc} = 0.067 * T_{otsu}
\end{align}

We consider that the gradient pixels values of the background, which are divided by the Otsu method, vary in a normal curve probability distribution \cite{patel1996handbook} form. So we can choose the mean value of the distribution as our gradient threshold. It seems appropriate to choose this value as gradient threshold because it will eliminate noise caused by small background changes, or small features in the image. We consider that anchors should be at the margins of the distribution, as peaks of the gradient map mountains, so we choose the anchor threshold as $6.7$ percent of Otsu threshold.

Another modification we proposed, that resulted from experiments, is to set the value of $1$ to the scan interval. From our observation, varying this parameter does not bring actual benefits in this new concept.

In Algorithm \ref{algorithm:ed_mod}, the proposed version of ED algorithm is described. If we look at the necessary input parameters, we can see that if we use this version we just have to set the smoothing kernel size.

We have chosen the weights for gradient and anchor thresholds from Otsu value using the assumption detailed in this section but other values can be experimented. With our proposed modification to the ED algorithm, we consider that the tuning phase of the algorithm is significantly reduced.

\section{ED simulation results}
\label{Section:ED_results}

All the simulation are done using EECVF - End-to-End Computer Vision Framework- \cite{eecvf, orhei2021end}, an open-source solution based on Python programming language, by running  the module $main\_modified\_ed\_algorithm$.

To find the optimal parameters for the best results, we vary the parameters as following: the Gaussian kernel size -$GK$- in the range of $3 \rightarrow 9$ using a step of $2$, the gradient threshold -$GT$- in range of $10 \rightarrow 150$ with a step of $10$, the anchor threshold -$TA$- in the following range $10 \rightarrow 60$ with a step of $10$ and the scan interval -$SI$- in the range $1 \rightarrow 5$.

We observe, in Figure \ref{fig:sobel_fine_tuning_natural}, that we obtain the best results using the following parameters: gradient threshold value of $50$, anchor threshold value of $10$, Gaussian kernel size of $9$ and scan interval of $1$. As stated in the introduction, to be able to find this set of parameters we had to run $1080$ variants. 

Using the found parameters, we changed the operator used in ED algorithm to observe the difference that we obtain, as we can see in Figure \ref{fig:ed_natural}. ED algorithm using operators like Sobel\cite{1973SobelOriginal}, Prewitt\cite{PrewittOriginal1970}, Kroon \cite{KroonOriginal2009}, Orhei \cite{2020Orhei} have similar results. At the other end we can clearly see that using Kayyali \cite{2000Kayyali}, Kirsch \cite{1971Kirschcomputer} or Kitchen\cite{Kitchen1989} produces worse results. We have chosen to vary the parameters with value of $10$ so we can obtain more accurate results.

The changing of the operator can have an important contribution to the resulted edge-map but they cannot overcome the wrong selection of parameters. If we look in  Figure \ref{fig:ed_natural}, when changing the operator, we do not see usually a big change, but if we look in Figure \ref{fig:sobel_fine_tuning_natural}, every change we did in one of the parameters produced a relevant change in metrics.

\begin{figure*}
\centering
\setlength{\tabcolsep}{0.5pt}
\begin{tabular}{ccccccccc} 
	    \includegraphics[width=0.11\textwidth]{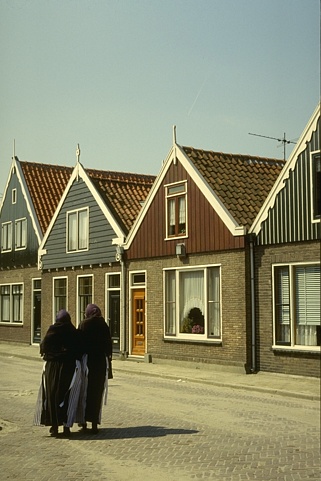} 
	&
        \includegraphics[width=0.11\textwidth]{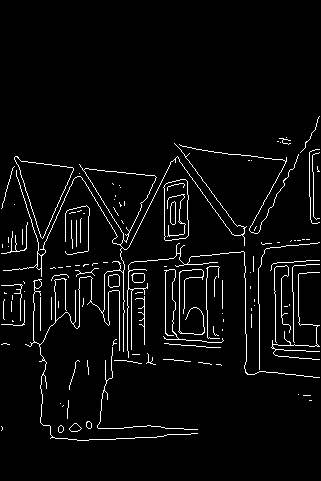} 
    & 
        \includegraphics[width=0.11\textwidth]{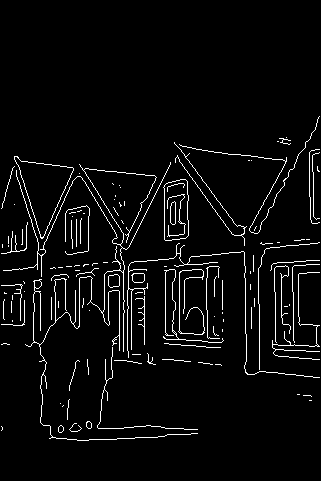}
    & 
        \includegraphics[width=0.11\textwidth]{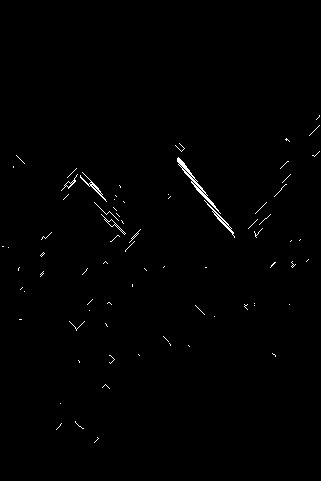}
    &
	    \includegraphics[width=0.11\textwidth]{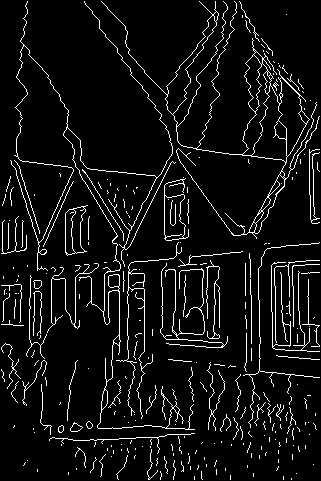} 
	&
        \includegraphics[width=0.11\textwidth]{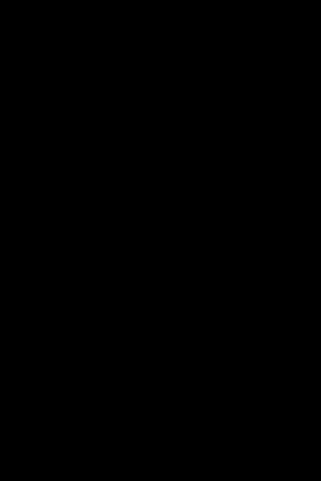} 
    & 
        \includegraphics[width=0.11\textwidth]{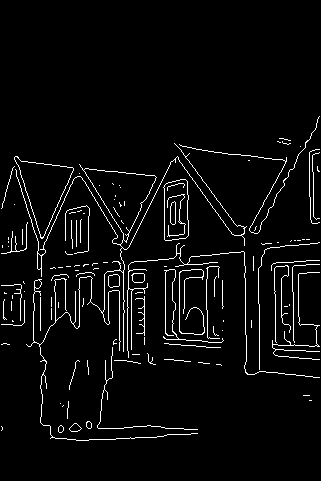}
    & 
        \includegraphics[width=0.11\textwidth]{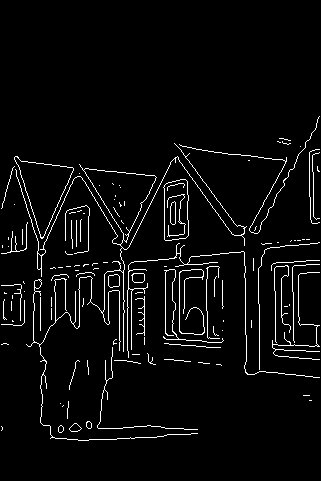}
    & 
        \includegraphics[width=0.11\textwidth]{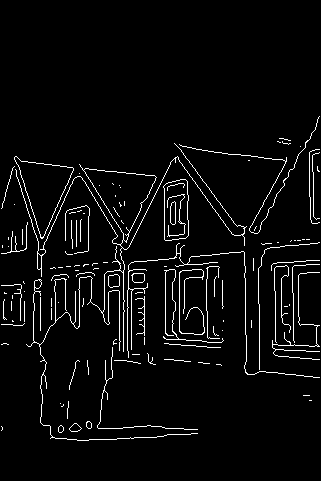}
    \\
	    \includegraphics[width=0.11\textwidth]{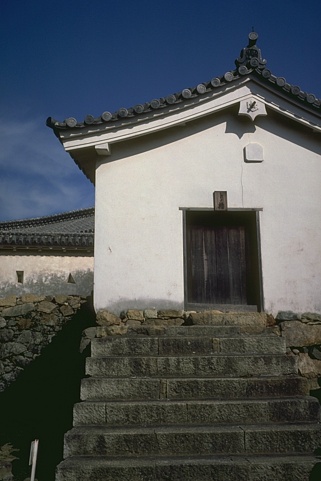} 
	&
        \includegraphics[width=0.11\textwidth]{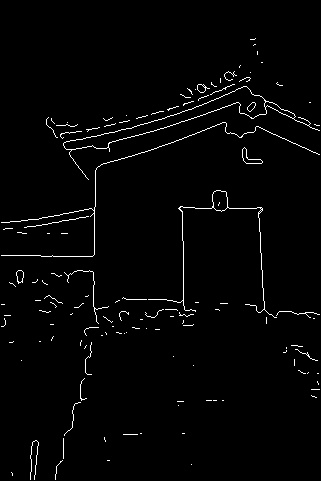} 
    & 
        \includegraphics[width=0.11\textwidth]{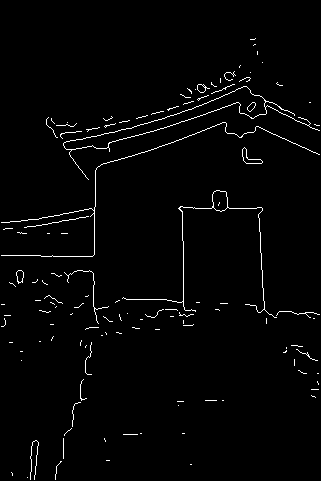}
    & 
        \includegraphics[width=0.11\textwidth]{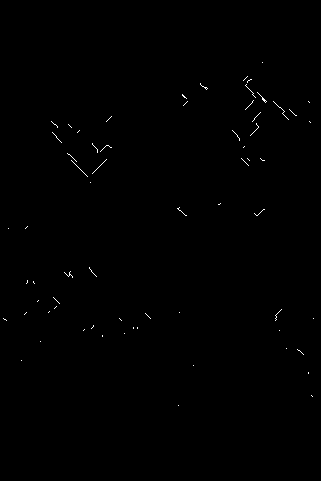}
    &
	    \includegraphics[width=0.11\textwidth]{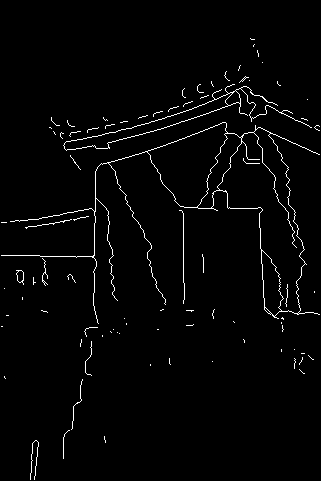} 
	&
        \includegraphics[width=0.11\textwidth]{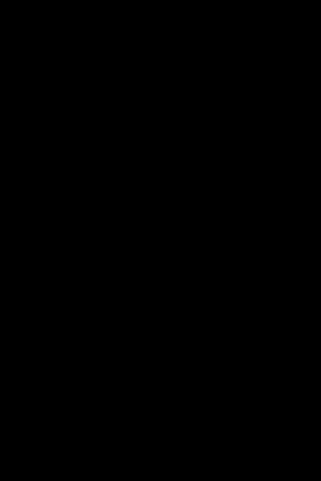} 
    & 
        \includegraphics[width=0.11\textwidth]{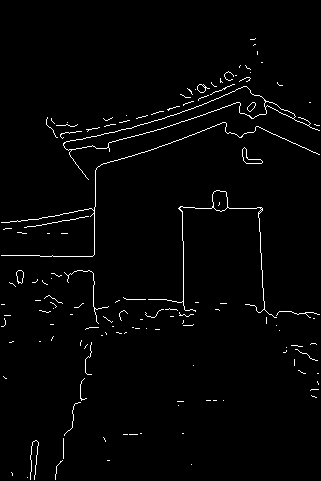}
    & 
        \includegraphics[width=0.11\textwidth]{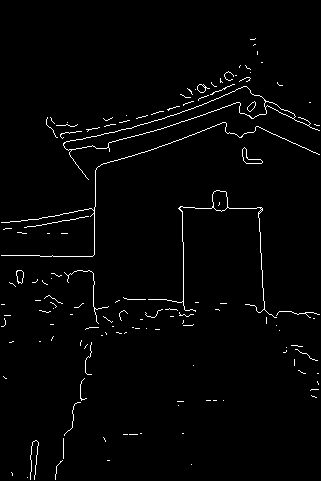}
    & 
        \includegraphics[width=0.11\textwidth]{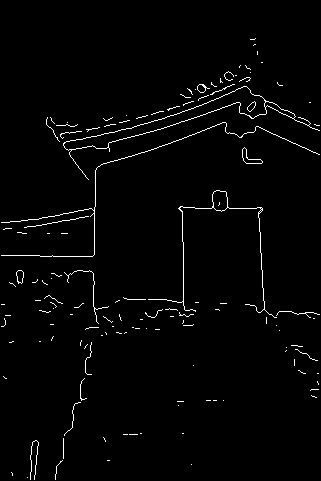}

\end{tabular}
\caption{Proposed ED algorithm visual results; Columns: original image, Sobel, Prewitt, Kirsch, Kitchen, Kayyali, Scharr, Kroon, Orhei;}
\label{fig:ed_img}
\end{figure*}

In Figure \ref{fig:ed_proposed} we present the statistical result and in Figure \ref{fig:ed_img} the visual results of the proposed ED algorithm. We can see that we obtain a slight decrease in the metrics, but we think it is an acceptable trade-off considering the amount of calculation we save by eliminating the fine-tuning phase.

Actually, for some cases we see a decrease in appeared artifacts (see Figure \ref{fig:ed_img}). To some extent, this is an expected outcome when switching from a threshold chosen by a global analysis of the image set to one that is image dependent.

        \begin{table}[!h]
          \centering
                \scalebox{0.9}
                {
                    \begin{tabular}{|l|cc|c|cc|c|}
                    \hline
                    {\bfseries Operator}    &\multicolumn{3}{c|}{\bfseries ED original} &\multicolumn{3}{c|}{\bfseries ED proposed}\\
                    \hline
                    \multicolumn{1}{|c|}{~}    &R   &P  &F1 &R   &P  &F1\\
                    \hline
                    \hline
                     Sobel          &0.721	&0.523	&0.606      &0.680	&0.515	&0.586\\
                    \hline
                     Prewitt        &0.720	&0.523	&0.606      &0.680	&0.515	&0.586\\
                    \hline
                     Kirsch         &0.139	&0.478	&0.216      &0.114	&0.474	&0.266\\
                    \hline
                     Kitchen        &0.724	&0.420	&0.531      &0.665	&0.414	&0.510\\
                    \hline
                     Kayyali        &0.000	&0.000	&0.000      &0.000	&0.000	&0.000\\
                    \hline
                     Scharr         &0.723	&0.522	&0.606      &0.681	&0.514	&0.586\\
                    \hline
                     Kroon          &0.723	&0.521	&0.606      &0.682	&0.513	&0.586\\
                    \hline
                     Orhei          &0.723	&0.522	&0.606      &0.681	&0.513	&0.586\\
                    \hline
                    \end{tabular}}
                \vspace{1.5pt}
            \caption{Comparison of ED results}
            \label{table:ed_results}
        \end{table}

We presented a comparison of the ED results, original versus proposed, for each operator, in Table \ref{table:ed_results}. The table reveals another important aspect: the proposed threshold finding method does not produce random results. Even if the results are lower, around 0.02 differences in $F1$ score, the order of the operators remain the same. 



\section{Conclusion}

In this paper we proposed a new Edge Drawing variant of the algorithm by adding an automated threshold choosing schema. This is an important aspect because it will reduce significantly the amount of preparation needed to use the algorithm. The proposed version can be used in different datasets or scenarios without the need of fine tuning first. 

The results we obtain with the proposed variant of ED are lower than the best variant found in our experiments. But to obtain the best variant result for one operator we needed more than a thousand tries, so in retrospect we can consider it an acceptable loss in metrics.

As future work, we consider using a more fine segmentation of the image to determine the thresholds by using Multi-Otsu thresholding \cite{liao2001fast}. Another aspect we wish to explore is the effect of dilated filters upon our proposed variant \cite{bogdan2020custom, orhei2020edge}.

\bibliographystyle{ieeetr}



\end{document}